\documentclass[conference,a4paper]{IEEEtran}
\usepackage{cite}
\usepackage{balance}
\usepackage{amsmath,amssymb,amsfonts}
\usepackage{algorithmic}
\usepackage{graphicx}
\usepackage{textcomp}
\usepackage{xcolor}
\def\BibTeX{{\rm B\kern-.05em{\sc i\kern-.025em b}\kern-.08em
    T\kern-.1667em\lower.7ex\hbox{E}\kern-.125emX}}

\usepackage[absolute,showboxes]{textpos}

\TPMargin{5pt}
\newcommand{\copyrightstatement}{
    \begin{textblock}{0.85}(0.08,0.90)    
         \noindent
         \footnotesize
Copyright 2019 IEEE. Published in the Digital Image Computing: Techniques and Applications, 2019 (DICTA 2019), 2-4 December 2019 in Perth, Australia. Personal use of this material is permitted. However, permission to reprint/republish this material for advertising or promotional purposes or for creating new collective works for resale or redistribution to servers or lists, or to reuse any copyrighted component of this work in other works, must be obtained from the IEEE. Contact: Manager, Copyrights and Permissions / IEEE Service Center / 445 Hoes Lane / P.O. Box 1331 / Piscataway, NJ 08855-1331, USA. Telephone: + Intl. 908-562-3966.
    \end{textblock}
}

\begin{document}

\title{Data-Efficient Classification of Birdcall Through Convolutional Neural Networks Transfer Learning\\
}

\author{
\IEEEauthorblockN{Dina B. Efremova}
\IEEEauthorblockA{\textit{Funbox Inc.} \\
Moscow, Russian Federation \\
dina.efremova85@gmail.com}
\and
\IEEEauthorblockN{Mangalam Sankupellay}
\IEEEauthorblockA{\textit{College of Science and Engineering} \\
\textit{James Cook University}\\
Townsville, Australia \\
mangalam.sankupellay@jcu.edu.au}
\and
\IEEEauthorblockN{Dmitry A. Konovalov}
\IEEEauthorblockA{\textit{College of Science and Engineering} \\
\textit{James Cook University}\\
Townsville, Australia \\
dmitry.konovalov@jcu.edu.au}
}

\maketitle
\copyrightstatement

\begin{abstract}
\boldmath
Deep learning Convolutional Neural Network (CNN) models are powerful classification models but require a large amount of training data. In niche domains such as bird acoustics, it is expensive and difficult to obtain a large number of training samples. One method of classifying data with a limited number of training samples is to employ transfer learning. In this research, we evaluated the effectiveness of birdcall classification using transfer learning from a larger base dataset (2814 samples in 46 classes) to a smaller target dataset (351 samples in 10 classes) using 
the ResNet-50  CNN. 
We obtained 79\% average validation accuracy on the target dataset 
in 5-fold cross-validation. The methodology of transfer learning from 
an ImageNet-trained CNN to a project-specific and a much smaller set of classes and images was extended to 
the domain of spectrogram images, where the base dataset 
effectively played the role of the ImageNet.
\end{abstract}


\section{Introduction}
Deep learning Convolutional Neural Network (CNN) models are powerful and popular classification architectures. CNN models have achieved the state-of-the-art results in the areas of image classification \cite{b1}, object detection \cite{b2}, face recognition \cite{b3}, and speech recognition \cite{b4} reaching high levels of accuracy \cite{b5}. In the area of image recognition, the success of CNN models is partially attributed to the availability of large-scale annotated datasets, e.g. ImageNet \cite{b6}. ImageNet is a comprehensive dataset with 1.2 million images in over 1,000 classes. CNN models, trained using ImageNet, learn through the high-level and layered hierarchy of image features. \par

While training data, for example, ImageNet, is relatively easily curated in the general image recognition domain, it is difficult to obtain a large amount of training data in niche areas such as medical imaging \cite{b6} or animal acoustics. For example, to obtain training data in animal acoustics, ecologists with expertise in specific animal calls have to manually listen to long duration (weeks or months-long) acoustic recordings and annotate these calls. This is a time-consuming, expensive endeavour prone to error due to human fatigue.\par

One method of classifying data with a limited number of training data is to employ transfer learning. Transfer learning is the reuse of a pre-trained model to solve a new problem \cite{b7} and is used to improve learning by transferring CNN connections, weights and biases, trained in one domain to a related or even different one \cite{b7}. Transfer learning is effective when there is a limited supply of target learning data due to the training data being rare, inaccessible, expensive, and/or time consuming to collect and label. Transfer learning has been successfully applied to medical image classification where the availability of training datasets is limited \cite{b6}.

\begin{figure*}[htp]
\centering
\includegraphics[scale=.40]{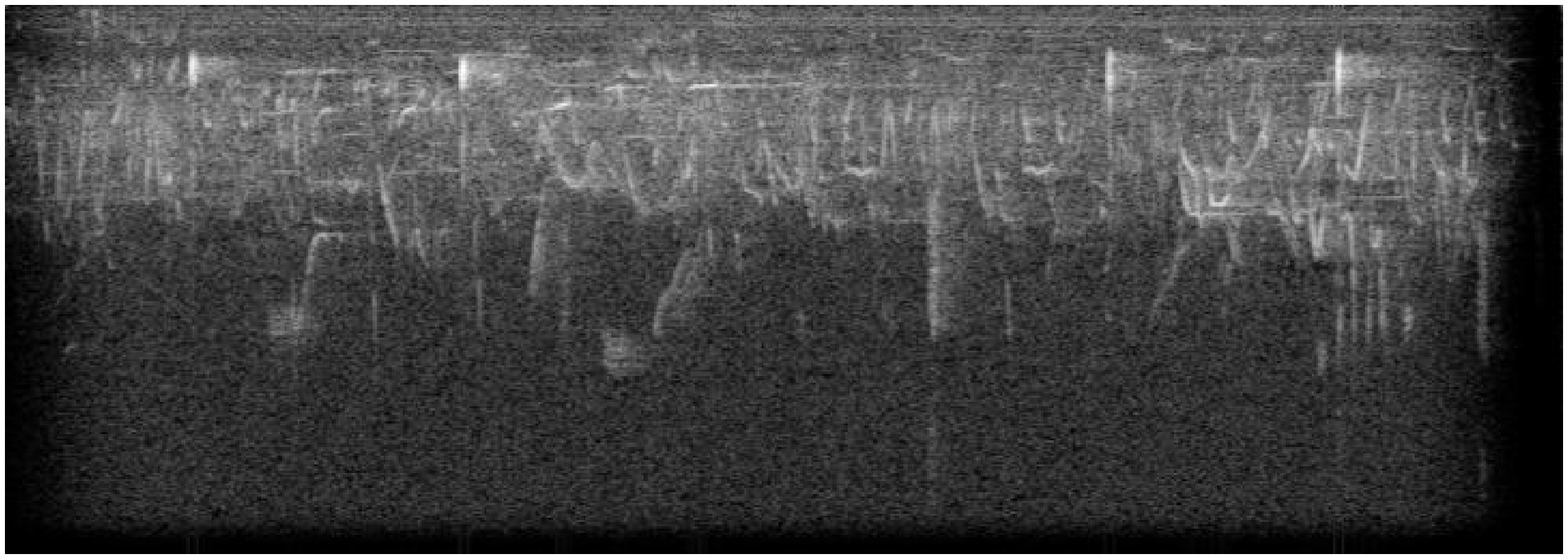} 
\includegraphics[scale=.40]{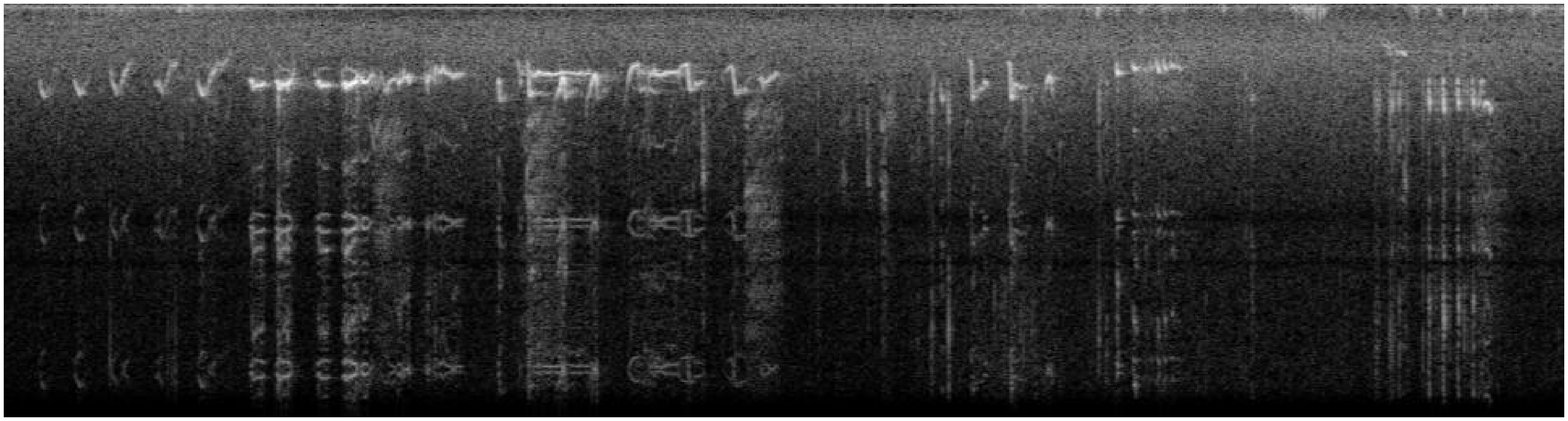}
\includegraphics[scale=.40]{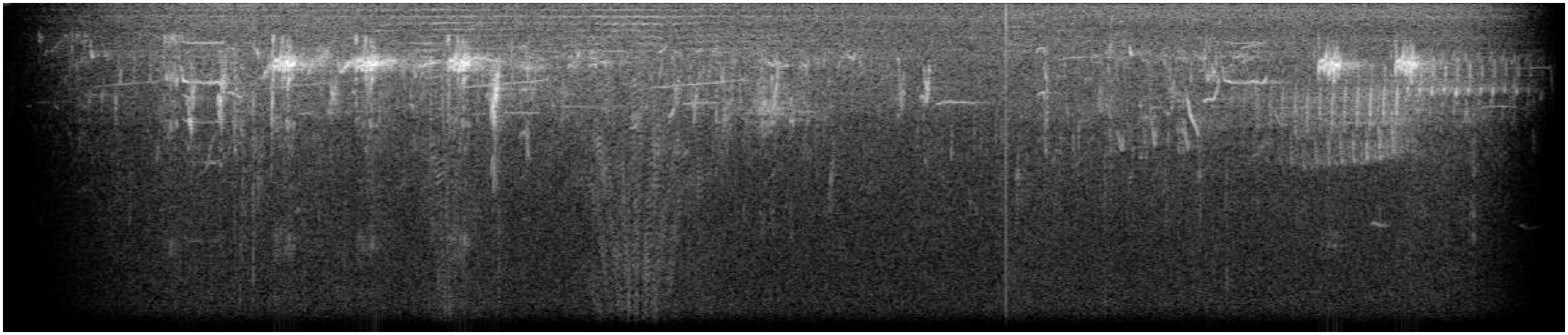} 
\caption{Three examples of {\em Acanthagenys rufogularis} birdcalls from the target dataset.}
\label{fig:aca}
\end{figure*}

\subsection{Birdcalls in Acoustic Recording}
The application of transfer learning in CNN could be beneficial in the area of animal call classification in environmental acoustic recordings due to the difficulty in obtaining annotated training calls. In this study, 
we investigated the application of transfer learning in CNN to classify birdcalls from environmental acoustic recordings. 

Ecologists and environmental managers use acoustic recordings obtained using Autonomous Recordings Units (ARUs) for long term non-invasive passive environmental monitoring. ARUs can be deployed in the field for weeks or months on end, over large spaces, and with minimal maintenance time and effort. As such, ARUs are a popular tool used by ecologists to easily monitor natural environments while reducing costly and time-consuming repeated visits to field sites.	

Ecologists use the acoustic recordings captured through ARUs for different purposes such as monitoring overall environmental health \cite{b8}, biodiversity \cite{b9}, threatened species \cite{b10}, invasive species \cite{b11}, occupancy of animals \cite{b12}, and climate change \cite{b13}. Most commonly, ecologists identify and then count the number of specific animal calls in an acoustic recording as a method of monitoring environmental changes. \par

Birds are one of the most important groups of animals ecologists monitor through acoustic recordings, as birds are an important indicator of biodiversity. The number and diversity of bird species in an ecosystem directly reflect biodiversity, ecosystem health, and suitability of the habitat \cite{b14}. Monitoring birdcalls in the ecosystem provides vital information about changes in the environment itself \cite{b14}. \par

Even though ARUs are a popular tool capturing acoustic recordings for environmental monitoring, there is a bottleneck in processing these acoustic recordings to identify specific birdcalls. 
Many ecologists rely on manual and 
time-consuming methods of listening to the recordings, as automated methods and tools for birdcall detection in acoustic recordings are still not available. The task of automatic birdcall classification in acoustic recordings is impacted by \cite{b14}:
\begin{itemize}
\item large inter- and intra-species birdcall variability;
\item environmental noise overlapping with birdcalls; 
\item overlapping birdcalls, especially during dawn and dusk choruses;
\item birds generating incomplete, quick calls or long calls in different situations, for example, birds generate quick calls during breeding season as they are occupied by incubation and/or chick rearing;
\item varying power in vocalisation due to distance and angles of birdcalls from the ARU microphones.
\end{itemize}
Fig.~\ref{fig:aca} illustrates the challenge of classifying bird
species by their birdcalls (at least via spectrograms), where 
all three sound segments were expertly labelled to belong to 
the same bird species, 
{\em Acanthagenys rufogularis}.

Some tools, such as SoundID \cite{b15} and Raven Pro \cite{b16}, use a semi-automated approach, but these tools require users to have considerable knowledge in signal processing making their use impractical to end users like ecologists. In addition, these tools require high calibration time as the recognisers are tailored (``{\em handcrafted}'') for specific birdcalls and do not generalise to other birdcalls \cite{b14}. Due to the challenges associated with birdcall classification, specifically the high variability in birdcalls, Machine Learning (ML) based automated birdcall classification is favoured because of the ability of ML algorithms to accommodate a high variability in birdcalls. 

Most ML approaches in animal call classification take their lead from automated speech recognition by virtue of the commonalities between human speech and birdcalls. These ML approaches include supervised neural networks (including deep learning neural networks) \cite{b17, b18, b19, b20,b21}, unsupervised neural networks \cite{b22}, support vector machines \cite{b23, b24, b25}, decision trees \cite{b26, b27}, random forests \cite{b28, b29}, and hidden markov model \cite{b30, b31, b32, b33, b34}. Despite the significant amount of research into the automated classification of birdcalls, there is not yet an adequate method for field recordings due to the challenges associated with birdcall classification, such as the high variability in calls. \par

Currently, supervised deep learning methods have gained popularity for automatic call classification in acoustic recordings. In the LifeCLEF Bird (Audio) Identification Task 2016/2017 algorithm benchmarking competition, the top algorithms were a variation of fully supervised deep learning CNN architecture \cite{b35, b36}. However, CNN models are heavily reliant on a large number of labelled samples, using experts to obtain such a large number of labelled records in acoustics is an expensive and time-consuming endeavour.\par

Yet, in an acoustic monitoring environment, it is relatively easy for ecologists to label a small number of animal calls focusing on the animal calls that they are interested in for a specific project or study. 
In addition, there is an abundance of annotated audio datasets with non-bird animal calls and calls from non-project specific birds that can be utilised. Given this scenario, transfer learning is a suitable technique to explore for birdcall classification.

Transfer learning is a method where a model developed for one task is reused/repurposed for a second related task. The first model is used as the starting point for the second task. Transfer learning is useful and important in deep learning given a large amount of data 
required to train a CNN model from scratch. In transfer learning, a source model is selected firstly. The source model is a pre-trained model that is trained on large and challenging datasets. The source model is then used as the starting point for the task of interest. In this, it may involve only using parts of the model or the whole model depending on the task of interest. The source model is fine adapted for the task at hand by fine-tuning the source model based on input-output pairs of the task of interests. 

Inspired by the success of CNN for birdcall classification in the LifeCLEF Bird (Audio) competition, in this research, we investigated the application of CNN transfer learning for birdcall classification using a relatively small number of training samples. 
Within the image classification domain, it is commonly accepted that 
the transfer learning method should be applied by retraining and/or fine-tuning an ImageNet-trained CNN using only project-specific images.
An alternative and not recommended approach would be to
add the project images into the pool of ImageNet images and 
retrain the CNN to classify the project classes as well as the 1,000 ImageNet classes at the same time.

Our main contribution was both practical and methodological.
In this study, we demonstrated 
how an ImageNet-like ``{\em SoundNet}'' collection of spectrograms
could be constructed
first and used to train a CNN. 
Then the SoundNet-trained CNN could be fine-tuned to classify a much 
smaller dataset of project-specific spectrograms. Therefore, the highly successful image-domain transfer learning approach could be replicated in nearly identical fashion for the sound spectrograms and used with confidence in future sounds classification studies.  

\begin{figure}[htbp]
\centerline{
    \includegraphics[width=0.49\textwidth]{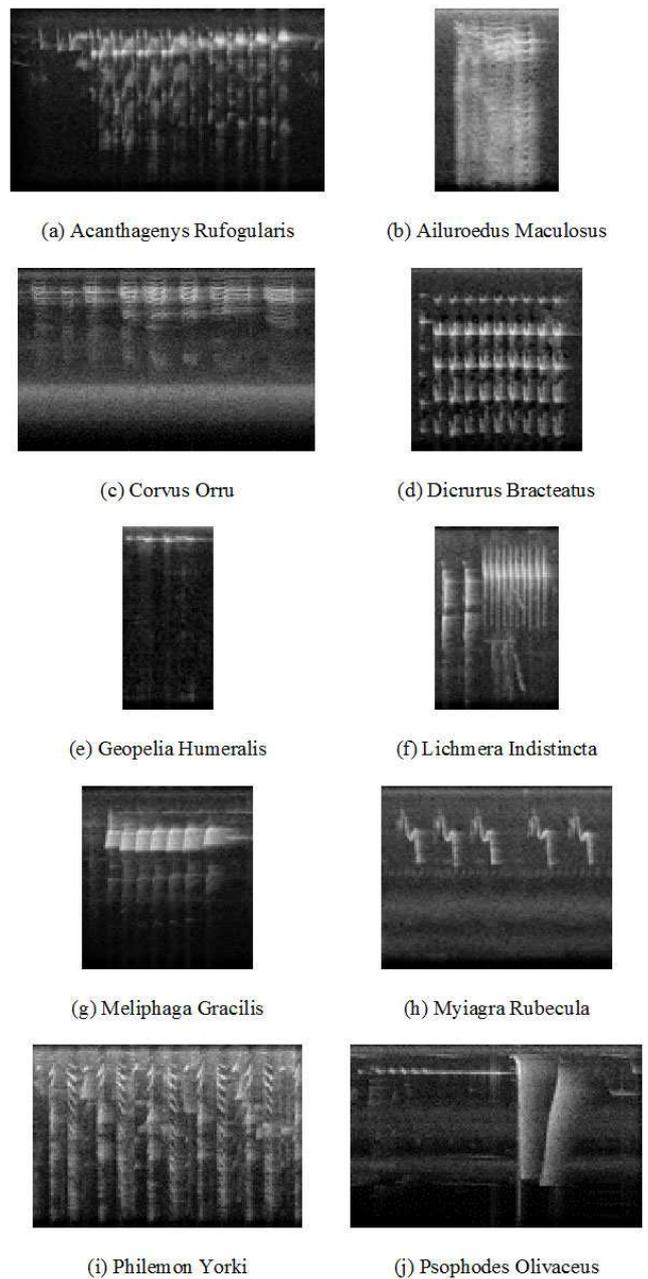}
}
\caption{Sample spectrograms of birdcalls from the target dataset.}
\label{fig:new-fig2}
\end{figure}

\section{Materials and methods}

\subsection{Dataset}
In this research, we used three different datasets to investigate the application of CNN transfer learning for birdcall classification as follows.

\subsubsection{Base ``SoundNet'' Dataset} 
Using the ImageNet as an analogy, 
in this study, a dataset developed by Nanni {\em et al.} \cite{b24}, from the Xeno-Canto site \cite{b37}, was selected as a base 
``{\em SoundNet}'' dataset.
It contained birdcalls recorded within a radius of up to 250 km from the city of Curitiba, in the South of Brazil. 
The dataset was publicly available, and it was a subset of Xeno-Canto set used in the BirdCELF challenges. 
Nanni {\em et al.} \cite{b24} removed all bird species with less than 10 samples. After these filters, 2814 audio samples representing 46 bird species remained in the dataset and were made available online
\footnote{https://bit.ly/2lLmcSW}. 22.05KHz was the sample rate of the audio files which were converted to spectrograms and made publicly available
\footnote{https://github.com/dmitryako/bs46spectrograms or https://bit.ly/2kCcUZs}.

\subsubsection{Target Dataset} 
The project target dataset used for 
transfer learning was a dataset developed from the Xeno-Canto site. 
The target dataset had birdcalls of 10 bird species common in 
the authors' home state of Queensland, Australia, and where at least 20 manually annotated (and with high confidence score) 
records existed at the Xeno-Canto site. 
This dataset had 351 audio samples representing 10 bird species 
(different from the base dataset's 46 bird species)
and it was 
made available
\footnote{https://github.com/dmitryako/aus10spectrograms or https://bit.ly/2k6sAnq}. 
The sample rate of the audio files was 41KHz.

\subsubsection{Negative Dataset} In addition to the base SoundNet and 
target datasets, the CNN model was trained using a negative dataset that was similar to the base and target datasets but from a different domain. For this purpose, a publicly available dataset \cite{b38} was used. The dataset had 16,930 sound instances of 243 environmental sounds, which were known 
not to be birdcalls. 

\subsection{Spectrograms}
The birdcalls and sounds in the base SoundNet, 
target, and negative datasets were converted into spectrogram images where the spectrum of frequencies (vertical y-axis, Hz) 
varied according to time (horizontal x-axis, seconds). 
The intensity of each pixel represents the frequency amplitude of the 
birdcall at a particular time. 
Since we worked with different quality sounds, for consistency, every sound recording was resampled to 22.05~KHz. 
The following spectrogram procedure was developed by experimenting
with different options to achieve visually expressive images, 
see examples in Figs.~\ref{fig:aca} and \ref{fig:new-fig2}.
Spectrograms were calculated using Fast Fourier Transformation (FFT) with a Hamming window with a frame length of $256 \times 4=1024$ samples and 
$(256-32) \times 4=896$ samples (87.5\%) overlap between subsequent frames. Intensities $S$ 
of the FFT-spectrograms were normalised to the same maximum value of 
$1 \times 10^8$ 
and then converted to the dB scale via $y=log(1+S)$. 
Due to the 1024-base FFT, all resulting images had 513 rows and a variable number of columns (i.e. different time durations of the original sound recordings). 
After extensive experimentation, it was found that the spectrograms could be proportionally downsized to have 256 rows, that made them more closely comparable with the standard image sizes used to train and test the modern ImageNet-trained CNN models. Then we normalised the images from 0-255 grayscale spectrogram to the [0,1]-range values. 

Note that the examples in Fig.~\ref{fig:new-fig2} were only one of many possible birdcalls for each species, while Fig.~\ref{fig:aca} 
depicts more realistic and much wider variations of birdcall patterns 
within the same species.

\begin{figure}[htbp]
\centering{\includegraphics[scale=0.7]{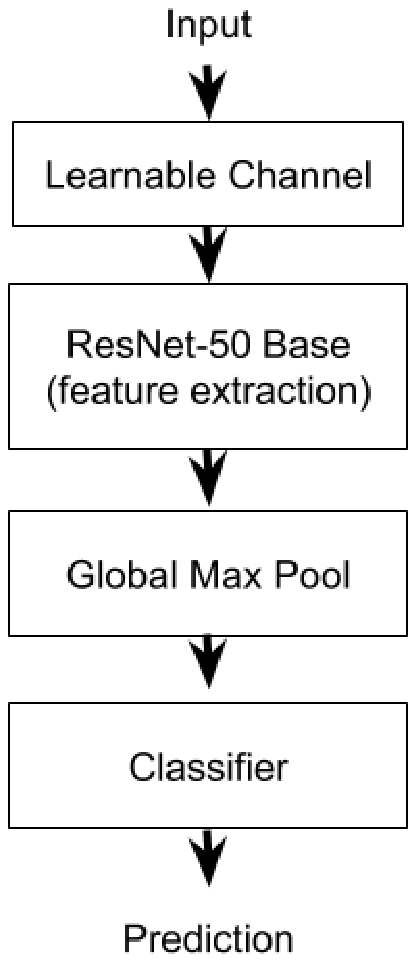}}
\caption{Model of the modified ResNet-50.}
\label{fig:resnet50}
\end{figure}

\subsection{Convolutional Neural Network Model}
The focus of this study was to verify the ImageNet-like 
transfer learning workflow, rather than to invent a better sound classification CNN.
Therefore,
we used well-established ResNet-50 baseline CNN, 
a 50 layer deep CNN architecture, to classify birdcalls. 
ResNet-50 was 
the first deep CNN architecture that utilised residual learning \cite{b2}. 
ResNet-50 has been successful in increasing accuracy in computer vision benchmarking challenges winning first prize in the ImageNet Large Scale Visual Recognition Challenge 2015 (ILSVRC, 2015) \cite{b40} and the Microsoft Common Objects in Context 2015 competition \cite{b2, b40}. The ResNet-50 model was trained on 1.28 million training images in 1,000 classes and reached an average of 5.25\% in top-5 errors \cite{b2}. In addition, the ResNet-50 model achieved 62\% accuracy in classifying 46 different bird species \cite{b41}.\par
We modified the ResNet-50 model for the classification of 
birdcalls as follows 
(Fig.~\ref{fig:resnet50}):
\begin{itemize}
\item
{\em A learnable channel} was added between the base ImageNet-trained ResNet-50 model and the input grayscale image (spectrogram) to convert the single-channel grayscale spectrogram for the expected by RetNet-50 
3-channel RGB image; 
\item After discarding the ImageNet classifier layer in the original ResNet-50,
a {\em global max-pooling} layer was added, followed by a 0.5 probability dropout layer to convert the last 2-dimensional (with
2048-channels) heatmap output of ResNet-50 into a 2048 feature vector;
\item The required classifications were achieved by 
adding a fully connected sigmoid-activated layer, \textit{classifier layer}, to accommodate the number of classes in either the base or 
target datasets (details in next section).
\end{itemize}

\subsection{ResNet-50 Base Dataset Training}
We used the ResNet-50 model that was available in the high-level neural network Application Programming Interface (API) of Keras \cite{b42} 
with the ML Python package, TensorFlow backend \cite{b43}. This model was trained to recognise the 1,000 different ImageNet \cite{b44} object classes. The original ImageNet-trained architecture was modified to classify 47 classes (46-class birdcall base dataset + 1 negative class sound dataset) by removing its 1,000-class top, adding the global 2D max pooling, 0.5 dropout, and a 47-neuron fully-connected layer. Specifically, the training spectrograms were randomly cropped to have 256 rows and 256 columns. The network then accepted a 
$256 \times 256 \times 1$ input image 
where the grayscale spectrogram image was converted into the three colour channels expected by the ResNet CNN via a trainable $1 \times 1$ convolution layer. After removing the ImageNet 1,000 classification layers, 
the ResNet-50 network outputs had the 
$8 \times 8 \times 2048$ shape, where 2048 was the number of extracted features for each $8 \times 8$ spatial location. 
The spatial max pooling layer was used to convert the fully-convolutional 
$8 \times 8 \times 2048$ output to the 2048 feature vector which was then densely connected (via the 0.5 dropout) to the final 47-classifier layer. A sigmoid activation function was used in the classification layer because, in practice, multiple birdcalls could be present in the same image. 
Hence, each class-specific sigmoid-activated neuron could independently detect a birdcall it was trained for in a given spectrogram. 

Prior to training, the ResNet-50 model was loaded with the corresponding ImageNet-trained weights available within Keras. In fact, this was the first knowledge transfer event of this study; that is, transferring the ImageNet domain of everyday images to the domain of sound spectrograms. Even for the cross-domain transfer, it was still more accurate and faster to train the ResNet-50 model with ImageNet-trained weights than to train a randomly initialised ResNet-50 model \cite{b45}. For the newly created gray-to-RGB conversion and 47-neuron fully-connected layers, the weights were initialised using uniform random distribution \cite{b46}. 
For training, the binary cross-entropy loss function was class-weighted. 
All not-ResNet-50 additional trainable weights were regularised by the 
$1 \times 10^{-5}$ weight decay.

To train the ResNet-50 model, the Adam \cite{b47} optimizer was used. The initial learning rate ($lr$) was set to $lr = 1 \times 10^{-5}$, 
which was relatively low to allow the ImageNet-trained weights to adjust gradually. 
It was then successively halved every time the validation loss did not decrease after 10 epochs, where the validation loss refers to 
the loss computed on the validation subset of images. 
While training, the model with the smallest running validation loss was continuously saved in order to restart the training after an abortion. The training was performed in batches of eight spectrograms and aborted if the validation loss did not decrease after 32 epochs. In such cases, the training cycle was repeated three more times with the initial learning rates scaled down by 0.9 at each restart.

All 2814 labelled spectrograms from the base SoundNet dataset were randomly partitioned into a 80\% and 20\% split of training and validation subsets, respectively, to monitor the training process and to estimate the predictive accuracy of the CNN. In addition, 1407 samples from the negative dataset were selected randomly for each epoch of training and validation. 
Before the $256 \times 256$ random crop, the spectrogram images were randomly scaled vertically and horizontally 
within the -10\% to 10\% range 
to account for the variability in birdcalls. After the crop, random uniform [0,25]-range noise was added at each pixel. And finally, the gray values were scaled to a minimum of zero and a maximum of one per image. Note that while the training images were randomly scaled and noise-added, the validation images were only randomly cropped and the [0,1]-range normalised.

\begin{figure}[htbp]
\centering
\includegraphics[width=0.47\textwidth]{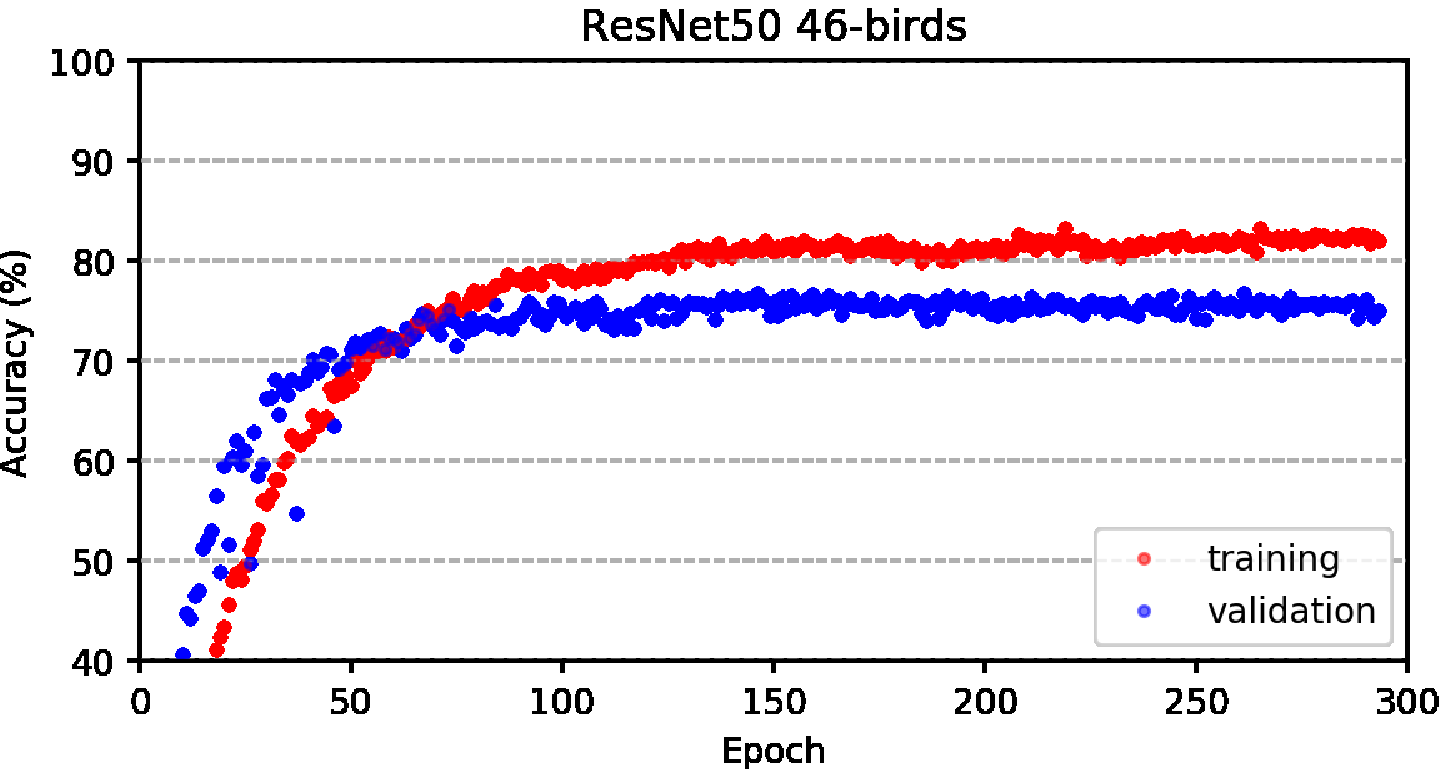}
\includegraphics[width=0.47\textwidth]{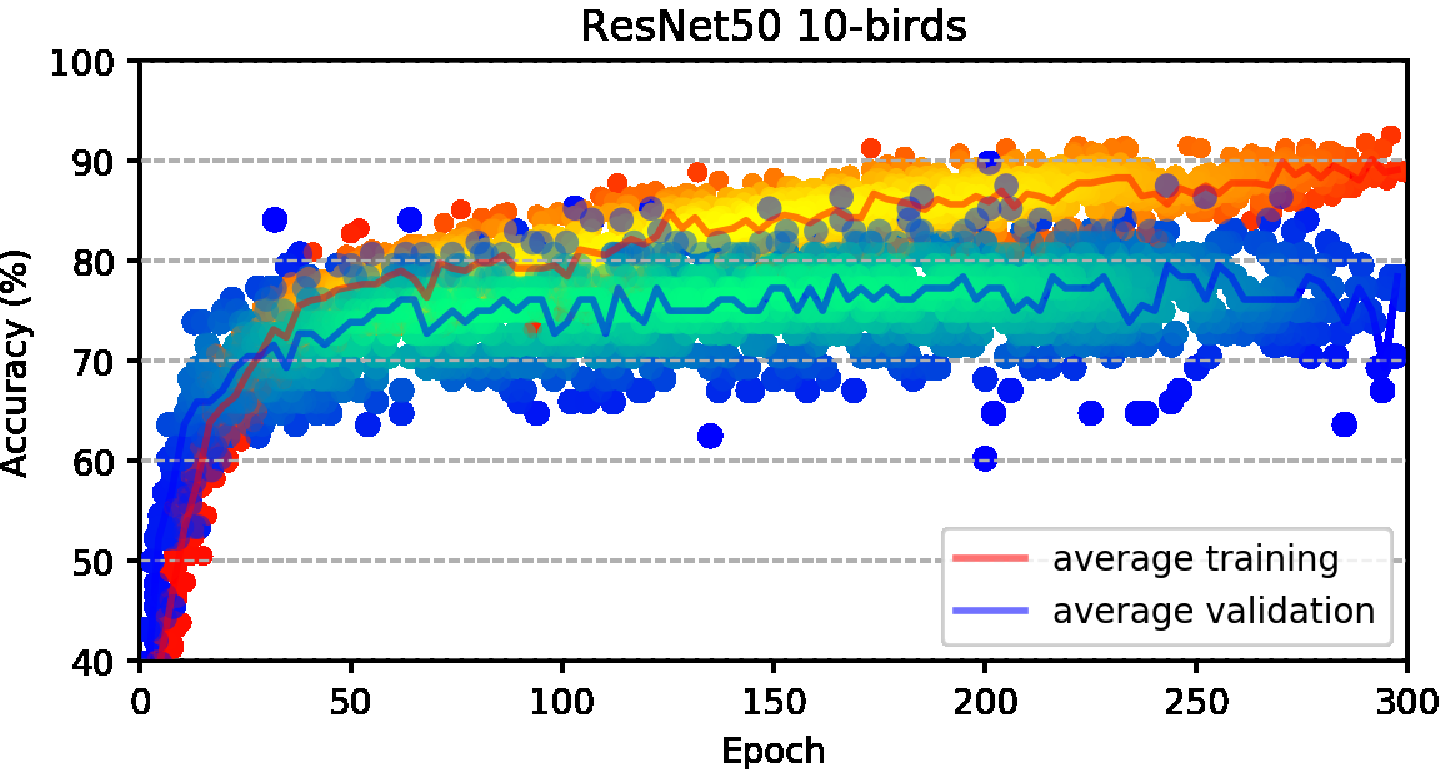}
\caption{Models' training and validation accuracy on: 
Single train-validation split for the base dataset (top); 
Five different train-validation splits for the target dataset (bottom)}
\label{fig:training}
\end{figure}

\subsection{ResNet-50 Target Dataset Training}
After training the ResNet-50 model with the 46-bird base 
``SoundNet'' dataset,
to transfer learning from the base dataset to the target 10-bird dataset, the ResNet-50 was modified to classify 11 classes (10-class birdcall base dataset + 1 negative class sound dataset). This was achieved by replacing the last densely-connected 47-neuron layer with a 11-neuron fully-connected layer. The training pipeline remained the same as per the preceding 47-class case; that is, the class-weighted binary cross-entropy loss function was used for training. Then ResNet-50 was trained with all 351 labelled spectrograms from the target dataset, which were randomly partitioned into a 72\% (i.e. 80\% of 90\%), 18\% (i.e. 20\% of 90\%), and 10\% split of training, validation, and testing subsets, respectively, 
to monitor the training process and to estimate the predictive accuracy of the CNN. In addition, 175 samples from the negative dataset were selected randomly for each epoch of training. 
Random five-fold cross-validation was performed: the complete training (from the 46-bird pretrained ResNet-50) cycle was repeated five times, where a different random seed was used each time to select a different subset of training, validation, and test images.

\begin{figure*}[htbp]
\centerline{\includegraphics[scale=0.60]{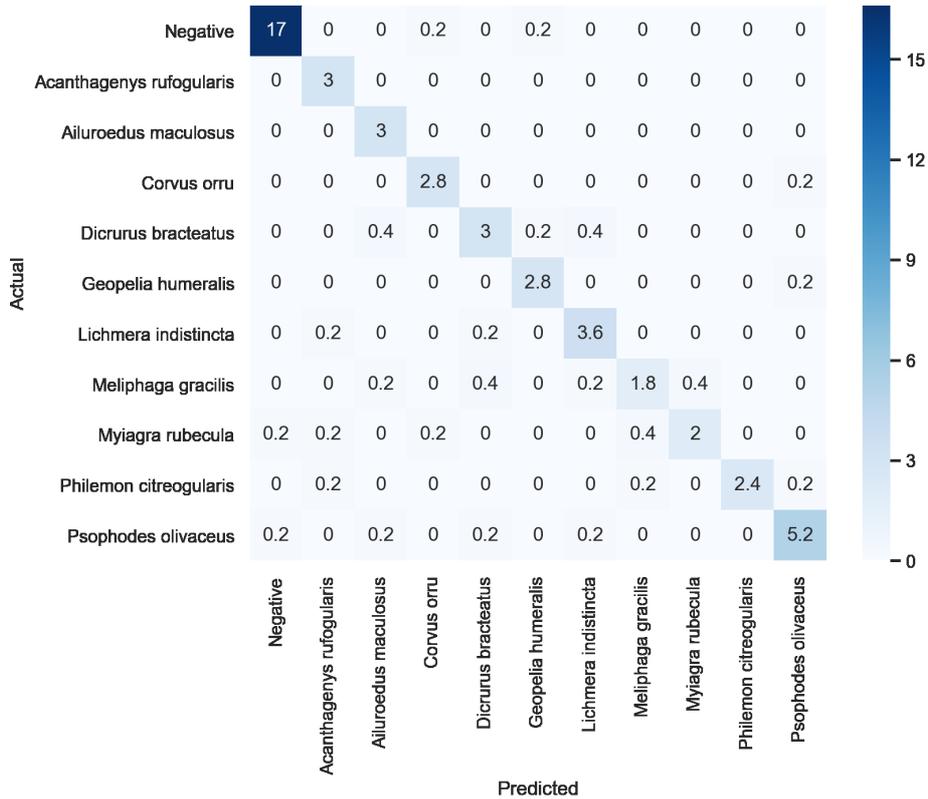}}
\caption{Five-fold averaged confusion matrix for 
test holdout subsets of the target dataset.}
\label{fig:confusion}
\end{figure*}

\section{RESULTS}
We used ResNet-50, a deep CNN architecture, for automated birdcall classification. We applied transfer learning from the base dataset of 46 different species of birds with a larger sample size (2814 samples) to a target dataset of 10 different species of birds with a 
much smaller sample size (351 samples).\par

\subsection{Spectrogram}
A total of 2814 spectrograms of birdcalls were 
generated for the base dataset and 351 spectrograms for the target dataset. 
Fig.~\ref{fig:new-fig2} shows sample spectrograms of 10 different bird species from the target dataset.

\subsection{ResNet-50 Transfer Learning}
Fig.~\ref{fig:training}(a) and Fig.~\ref{fig:training}(b) 
present the training process for the ResNet-50 model on both the base and target birdcall datasets, respectively. Lighter colours indicate higher density of points in Fig.~\ref{fig:training}(b). 
For both datasets, the ResNet-50 was trained on 
$256 (height) \times 256 (width)$ images randomly cropped from the spectrograms.\par
For the base dataset training, the network reached about 82\% training accuracy and 78\% validation accuracy. The accuracy began to plateau after 150 epochs. It took approximately 10 hours to train the ResNet-50 model on a Nvidia GTX 1080 Ti GPU. \par

With this transfer learning and further training, the network reached about 89\% training accuracy and 79\% validation accuracy for the target dataset. The accuracy began to plateau after 50 epochs. It took approximately 2 hours to train the ResNet-50 model on a Nvidia GTX 1080 Ti GPU.\par
The training accuracy in both instances exceeded 
the validation accuracy by only a small amount ($< \sim 9\%$). This was indicative of a network that was not underfitting or overfitting to the training data. Note that only the additional training noise, random rows and columns scaling, and the much larger negative dataset prevented the ResNet-50 model from drastically overfitting such a small target dataset (only 351 images for 10 birds). 

Fig.~\ref{fig:confusion} shows the confusion matrix of actual versus predicted classification of the testing samples of the target dataset (averaged over the five train/test cross validations). As expected, the negative class (non-birdcall class) had the highest correct classification. 
Among the target dataset birdcalls, class 10 (Fig.~\ref{fig:new-fig2}(j) {\em Psophodes olivaceus}) 
had the highest correct classification due to its very
distinct birdcall signature, 
while class 7 (Fig.~\ref{fig:new-fig2}(g) 
{\em Meliphaga gracilis}) had the lowest correct classification. 
For testing, each test image was converted to a series of 50\%-column overlapping $256 \times 256$ images, 
and then the maximum class prediction value (for each of the 11 classes) was used to assign the classification prediction of the test image. While only one bird species per image was assumed in this study, the same testing procedure could be used to extract multiple bird species from the same image in the future, e.g. by using an activation 
level threshold.

This was the first reported research on the application  of  CNN model in  birdcall  classification  utilising transfer  learning  from  a  larger  base  dataset  to  a  smaller target  dataset. There is no prior research (baseline) available to compare with.  

\section{DISCUSSION AND CONCLUSION}
In this study, we evaluated the application of transfer learning for the classification of birdcalls. We evaluated the application of transfer learning from a larger-base bird-sound dataset (2814 sounds) to a smaller target dataset (351 sounds) as it was difficult to obtain a 
large number of birdcalls for a specific bird species. In addition to the development of cross- and within-domain knowledge transfer procedures, 
we developed a new (at least for the sound domain) regularisation technique of using a much larger pool of negative examples, consisting of environmental sounds (non-birdcalls). 
The large variety of negative samples forced the training to focus on the birdcalls rather than on the non-bird surrounding sounds, which assisted in preventing the overfitting of the relatively small number of training samples by the high capacity ResNet-50 CNN. 
We used the deep CNN, ResNet-50, for feature extraction and classification due to ResNet-50's successful image classification in the ILSVRC 2015 and MS COCO 2015 competitions \cite{b40}. In addition, ResNet-50 has been successful in classifying birdcalls \cite{b41}.

Firstly, we trained the entire ResNet-50 with the larger base birdcall dataset (2814 samples) and a negative class of environmental sounds (16,930 samples). 
The validation accuracy of ResNet-50 reached 75\% and 
plateaued at 
around 150 epochs. 
This was more accurate and faster than in the previous work by Sankupellay and Konovalov \cite{b41} where the ResNet-50 validation accuracy was only 65\% at 
around 300 epochs. 
The 10\% improvement in validation accuracy and 
twice faster training speed were attributed to: 
\begin{itemize}
\item using the 
$256 \times 256$ training image sizes, which were 
closer to the intended use of ResNet-50, 
and where 512-row images were used in \cite{b41}; 
\item allowing the CNN architecture to automatically adjust its input via the gray-to-RGB trainable conversion layer, where the
typical learnt conversion weights were 
$\{0.55, -0.145, -0.54\}$ and zero biases; 
\item the regularisation via the much larger negative class dataset;
and 
\item using the maximum pool layer instead of the average pooling layer in \cite{b41},
which contributed around 2\% to the accuracy but not to the speed of training.
\end{itemize}

Then, we applied transfer learning from the larger base dataset to smaller target dataset (only 351 samples) by fine-tuning ResNet-50. Effectively, features extracted from the larger base dataset were utilised for the classification of the smaller target dataset.
In this research, we achieved 79\% validation classification accuracy with 
a data-efficient small number of birdcall samples.
\balance
\bibliographystyle{IEEEtran}
\bibliography{IEEEabrv,birds2019}

\begin{thebibliography}{10}
\providecommand{\url}[1]{#1}
\csname url@samestyle\endcsname
\providecommand{\newblock}{\relax}
\providecommand{\bibinfo}[2]{#2}
\providecommand{\BIBentrySTDinterwordspacing}{\spaceskip=0pt\relax}
\providecommand{\BIBentryALTinterwordstretchfactor}{4}
\providecommand{\BIBentryALTinterwordspacing}{\spaceskip=\fontdimen2\font plus
\BIBentryALTinterwordstretchfactor\fontdimen3\font minus
  \fontdimen4\font\relax}
\providecommand{\BIBforeignlanguage}[2]{{%
\expandafter\ifx\csname l@#1\endcsname\relax
\typeout{** WARNING: IEEEtran.bst: No hyphenation pattern has been}%
\typeout{** loaded for the language `#1'. Using the pattern for}%
\typeout{** the default language instead.}%
\else
\language=\csname l@#1\endcsname
\fi
#2}}
\providecommand{\BIBdecl}{\relax}
\BIBdecl

\bibitem{b1}
A.~Krizhevsky, I.~Sutskever, and G.~E. Hinton, ``Imagenet classification with
  deep convolutional neural networks,'' in \emph{Advances in Neural Information
  Processing Systems 25}, F.~Pereira, C.~J.~C. Burges, L.~Bottou, and K.~Q.
  Weinberger, Eds.\hskip 1em plus 0.5em minus 0.4em\relax Curran Associates,
  Inc., 2012, pp. 1097--1105.

\bibitem{b2}
K.~He, X.~Zhang, S.~Ren, and J.~Sun, ``Deep residual learning for image
  recognition,'' in \emph{2016 IEEE Conference on Computer Vision and Pattern
  Recognition (CVPR)}, 2016, pp. 770--778.

\bibitem{b3}
H.~Li, Z.~Lin, X.~Shen, J.~Brandt, and G.~Hua, ``A convolutional neural network
  cascade for face detection,'' in \emph{2015 IEEE Conference on Computer
  Vision and Pattern Recognition (CVPR)}, 2015, pp. 5325--5334.

\bibitem{b4}
S.~S. Farfade, M.~J. Saberian, and L.-J. Li, ``Multi-view face detection using
  deep convolutional neural networks,'' in \emph{Proceedings of the 5th ACM on
  International Conference on Multimedia Retrieval}, ser. ICMR '15.\hskip 1em
  plus 0.5em minus 0.4em\relax New York, NY, USA: ACM, 2015, pp. 643--650.

\bibitem{b5}
I.~Goodfellow, Y.~Bengio, and A.~Courville, \emph{Deep Learning}.\hskip 1em
  plus 0.5em minus 0.4em\relax MIT Press, 2016,
  \url{http://www.deeplearningbook.org}.

\bibitem{b6}
H.~Shin, H.~R. Roth, M.~Gao, L.~Lu, Z.~Xu, I.~Nogues, J.~Yao, D.~Mollura, and
  R.~M. Summers, ``Deep convolutional neural networks for computer-aided
  detection: Cnn architectures, dataset characteristics and transfer
  learning,'' \emph{IEEE Transactions on Medical Imaging}, vol.~35, pp.
  1285--1298, 2016.

\bibitem{b7}
S.~J. Pan and Q.~Yang, ``A survey on transfer learning,'' \emph{IEEE
  Transactions on Knowledge and Data Engineering}, vol.~22, pp. 1345--1359,
  2010.

\bibitem{b8}
E.~P. Kasten, S.~H. Gage, J.~Fox, and W.~Joo, ``The remote environmental
  assessment laboratory's acoustic library: An archive for studying soundscape
  ecology,'' \emph{Ecological Informatics}, vol.~12, pp. 50 -- 67, 2012.

\bibitem{b9}
A.~Gasc, J.~Sueur, S.~Pavoine, R.~Pellens, and P.~Grandcolas, ``Biodiversity
  sampling using a global acoustic approach: Contrasting sites with
  microendemics in new caledonia,'' \emph{PLOS ONE}, vol.~8, pp. 1--10, 2013.

\bibitem{b10}
M.~Campos-Cerqueira and T.~M. Aide, ``Improving distribution data of threatened
  species by combining acoustic monitoring and occupancy modelling,''
  \emph{Methods in Ecology and Evolution}, vol.~7, pp. 1340--1348, 2016.

\bibitem{b11}
W.~Hu, N.~Bulusu, C.~T. Chou, S.~Jha, A.~Taylor, and V.~N. Tran, ``Design and
  evaluation of a hybrid sensor network for cane toad monitoring,'' \emph{ACM
  Transactions on Sensor Networks (TOSN)}, vol.~5, pp. 1--28, 2009.

\bibitem{b12}
A.~K. Kalan, R.~Mundry, O.~J. Wagner, S.~Heinicke, C.~Boesch, and H.~S. Kuhl,
  ``Towards the automated detection and occupancy estimation of primates using
  passive acoustic monitoring,'' \emph{Ecological Indicators}, vol.~54, pp.
  217--226, 2015.

\bibitem{b13}
A.~Farina, \emph{Soundscape Ecology: Principles, Patterns, Methods and
  Applications}.\hskip 1em plus 0.5em minus 0.4em\relax Springer, 2014.

\bibitem{b14}
N.~Priyadarshani, S.~Marsland, and I.~Castro, ``Automated birdsong recognition
  in complex acoustic environments: a review,'' \emph{Journal of Avian
  Biology}, vol.~49, pp. 1--27, 2018.

\bibitem{b15}
\BIBentryALTinterwordspacing
N.~J. Boucher, ``{SoundID} version 2.0.0 documentation,'' 2014. [Online].
  Available: \url{https://bit.ly/2LZKlOY}
\BIBentrySTDinterwordspacing

\bibitem{b16}
\BIBentryALTinterwordspacing
{Bioacoustics Research Program, The Cornell Lab of Ornithology}, ``{Raven Pro:
  Interactive Sound Analysis Software (Version 1.5)},'' 2014. [Online].
  Available: \url{https://ravensoundsoftware.com/}
\BIBentrySTDinterwordspacing

\bibitem{b17}
J.~Cai, D.~Ee, B.~Pham, P.~Roe, and J.~Zhang, ``Sensor network for the
  monitoring of ecosystem: Bird species recognition,'' in \emph{2007 3rd
  International Conference on Intelligent Sensors, Sensor Networks and
  Information}, 2007, pp. 293--298.

\bibitem{b18}
M.~T. Lopes, L.~L. Gioppo, T.~T. Higushi, C.~A.~A. Kaestner, C.~N.~S. Jr., and
  A.~L. Koerich, ``Automatic bird species identification for large number of
  species,'' in \emph{2011 IEEE International Symposium on Multimedia}, 2011,
  pp. 117--122.

\bibitem{b19}
D.~Chakraborty, P.~Mukker, P.~Rajan, and A.~D. Dileep, ``Bird call
  identification using dynamic kernel based support vector machines and deep
  neural networks,'' in \emph{2016 15th IEEE International Conference on
  Machine Learning and Applications (ICMLA)}, 2016, pp. 280--285.

\bibitem{b20}
S.~Qi, Z.~Huang, Y.~Li, and S.~Shi, ``Audio recording device identification
  based on deep learning,'' in \emph{2016 IEEE International Conference on
  Signal and Image Processing (ICSIP)}, 2016, pp. 426--431.

\bibitem{b21}
A.~Sevilla and H.~Glotin, ``Audio bird classification with inception-v4
  extended with time and time-frequency attention mechanisms,'' in
  \emph{Working Notes of CLEF}, 2017.

\bibitem{b22}
D.~Stowell and M.~D. Plumbley, ``Automatic large-scale classification of bird
  sounds is strongly improved by unsupervised feature learning,'' \emph{PeerJ},
  vol.~2, pp. 1--31, 2014.

\bibitem{b23}
O.~Dufour, T.~Artieres, H.~Glotin, and P.~Giraudet, ``Clusterized mel filter
  cepstral coefficients and support vector machines for bird song
  identification,'' in \emph{Soundscape Semiotics - Localization and
  Categorization}, H.~Glotin, Ed.\hskip 1em plus 0.5em minus 0.4em\relax
  InTech, 2014, pp. 83--95.

\bibitem{b24}
L.~Nanni, Y.~M.~G. Costa, D.~R. Lucio, C.~N. Silla, and S.~Brahnam, ``Combining
  visual and acoustic features for bird species classification,'' in \emph{2016
  IEEE 28th International Conference on Tools with Artificial Intelligence
  (ICTAI)}, 2016, pp. 396--401.

\bibitem{b25}
L.~Nanni, Y.~Costa, D.~Lucio, C.~Silla, and S.~Brahnam, ``Combining visual and
  acoustic features for audio classification tasks,'' \emph{Pattern Recognition
  Letters}, vol.~88, pp. 49--56, 2017.

\bibitem{b26}
A.~Digby, M.~Towsey, B.~D. Bell, and P.~D. Teal, ``A practical comparison of
  manual and autonomous methods for acoustic monitoring,'' \emph{Methods in
  Ecology and Evolution}, vol.~4, pp. 675--683, 2013.

\bibitem{b27}
M.~Lasseck, ``Towards automatic large-scale identification of birds in audio
  recordings,'' in \emph{Experimental IR Meets Multilinguality, Multimodality,
  and Interaction}, J.~Mothe, J.~Savoy, J.~Kamps, K.~Pinel-Sauvagnat, G.~Jones,
  E.~San~Juan, L.~Capellato, and N.~Ferro, Eds.\hskip 1em plus 0.5em minus
  0.4em\relax Cham: Springer International Publishing, 2015, pp. 364--375.

\bibitem{b28}
I.~Potamitis, S.~Ntalampiras, O.~Jahn, and K.~Riede, ``Automatic bird sound
  detection in long real-field recordings: Applications and tools,''
  \emph{Applied Acoustics}, vol.~80, pp. 1--9, 2014.

\bibitem{b29}
G.~Fodor, ``The ninth annual {MLSP} competition: First place,'' in \emph{2013
  IEEE International Workshop on Machine Learning for Signal Processing
  (MLSP)}, 2013, pp. 1--2.

\bibitem{b30}
J.~A. Kogan and D.~Margoliash, ``Automated recognition of bird song elements
  from continuous recordings using dynamic time warping and hidden markov
  models: A comparative study,'' \emph{The Journal of the Acoustical Society of
  America}, vol. 103, pp. 2185--2196, 1998.

\bibitem{b31}
C.~Kwan, G.~Mei, X.~Zhao, Z.~Ren, R.~Xu, V.~Stanford, C.~Rochet, J.~Aube, and
  K.~C. Ho, ``Bird classification algorithms: theory and experimental
  results,'' in \emph{2004 IEEE International Conference on Acoustics, Speech,
  and Signal Processing}, vol.~5, 2004, pp. V--289.

\bibitem{b32}
\BIBentryALTinterwordspacing
V.~M. Trifa, ``A framework for bird songs detection, recognition and
  localization using acoustic sensor networks,'' 2006. [Online]. Available:
  \url{https://bit.ly/2sdgbyn}
\BIBentrySTDinterwordspacing

\bibitem{b33}
P.~Jancovic and M.~K{\"o}k{\"u}er, ``Acoustic recognition of multiple bird
  species based on penalized maximum likelihood,'' \emph{IEEE Signal Processing
  Letters}, vol.~22, pp. 1585--1589, 2015.

\bibitem{b34}
------, ``Recognition of multiple bird species based on penalised maximum
  likelihood and hmm-based modelling of individual vocalisation elements,'' in
  \emph{INTERSPEECH}, 2016, pp. 2612--2616.

\bibitem{b35}
A.~Joly, H.~Go{\"e}au, H.~Glotin, C.~Spampinato, P.~Bonnet, W.-P. Vellinga,
  J.~Champ, R.~Planqu{\'e}, S.~Palazzo, and H.~M{\"u}ller, ``{LifeCLEF 2016}:
  Multimedia life species identification challenges,'' in \emph{Experimental IR
  Meets Multilinguality, Multimodality, and Interaction}, N.~Fuhr, P.~Quaresma,
  T.~Gon{\c{c}}alves, B.~Larsen, K.~Balog, C.~Macdonald, L.~Cappellato, and
  N.~Ferro, Eds.\hskip 1em plus 0.5em minus 0.4em\relax Cham: Springer
  International Publishing, 2016, pp. 286--310.

\bibitem{b36}
A.~Joly, H.~Go{\"e}au, H.~Glotin, C.~Spampinato, P.~Bonnet, W.-P. Vellinga,
  J.-C. Lombardo, R.~Planqu{\'e}, S.~Palazzo, and H.~M{\"u}ller, ``{LifeCLEF
  2017 Lab Overview}: Multimedia species identification challenges,'' in
  \emph{Experimental IR Meets Multilinguality, Multimodality, and Interaction},
  G.~J. Jones, S.~Lawless, J.~Gonzalo, L.~Kelly, L.~Goeuriot, T.~Mandl,
  L.~Cappellato, and N.~Ferro, Eds.\hskip 1em plus 0.5em minus 0.4em\relax
  Cham: Springer International Publishing, 2017, pp. 255--274.

\bibitem{b37}
\BIBentryALTinterwordspacing
{Xeno-canto foundation}, 2018. [Online]. Available:
  \url{https://www.xeno-canto.org}
\BIBentrySTDinterwordspacing

\bibitem{b38}
W.~Han, E.~Coutinho, H.~Ruan, H.~Li, B.~Schuller, X.~Yu, and X.~Zhu,
  ``Semi-supervised active learning for sound classification in hybrid learning
  environments,'' \emph{PLOS ONE}, vol.~11, pp. 1--23, 2016.

\bibitem{b40}
\BIBentryALTinterwordspacing
{ImageNet Large Scale Visual Recognition Challenge 2015 (ILSVRC2015)},
  ``Results,'' 2015. [Online]. Available: \url{https://bit.ly/21Us87z}
\BIBentrySTDinterwordspacing

\bibitem{b41}
\BIBentryALTinterwordspacing
M.~Sankupellay and D.~A. Konovalov, ``Bird call recognition using deep
  convolutional neural network, {ResNet-50},'' in \emph{ACOUSTICS}, 2018.
  [Online]. Available: \url{https://bit.ly/2lDcdPQ}
\BIBentrySTDinterwordspacing

\bibitem{b42}
\BIBentryALTinterwordspacing
F.~Chollet \emph{et~al.}, ``Keras: The python deep learning library,'' 2015.
  [Online]. Available: \url{https://keras.io/}
\BIBentrySTDinterwordspacing

\bibitem{b43}
\BIBentryALTinterwordspacing
M.~Abadi \emph{et~al.}, ``{TensorFlow}: Large-scale machine learning on
  heterogeneous systems,'' 2015. [Online]. Available:
  \url{http://tensorflow.org/}
\BIBentrySTDinterwordspacing

\bibitem{b44}
O.~Russakovsky, J.~Deng, H.~Su, J.~Krause, S.~Satheesh, S.~Ma, Z.~Huang,
  A.~Karpathy, A.~Khosla, M.~Bernstein, A.~C. Berg, and L.~Fei-Fei, ``Imagenet
  large scale visual recognition challenge,'' \emph{International Journal of
  Computer Vision}, vol. 115, pp. 211--252, 2015.

\bibitem{b45}
M.~Oquab, L.~Bottou, I.~Laptev, and J.~Sivic, ``Learning and transferring
  mid-level image representations using convolutional neural networks,'' in
  \emph{2014 IEEE Conference on Computer Vision and Pattern Recognition}.\hskip
  1em plus 0.5em minus 0.4em\relax IEEE, 2014, pp. 1717--1724.

\bibitem{b46}
X.~Glorot and Y.~Bengio, ``Understanding the difficulty of training deep
  feedforward neural networks,'' in \emph{Proceedings of the Thirteenth
  International Conference on Artificial Intelligence and Statistics}, ser.
  Proceedings of Machine Learning Research, Y.~W. Teh and M.~Titterington,
  Eds., vol.~9.\hskip 1em plus 0.5em minus 0.4em\relax PMLR, 2010, pp.
  249--256.

\bibitem{b47}
D.~P. Kingma and J.~Ba, ``Adam: {A} method for stochastic optimization,'' in
  \emph{3rd International Conference on Learning Representations (ICLR2015)},
  2015.

\end{thebibliography}

\end{document}